\documentclass[10pt,twocolumn]{article}
\pdfoutput=1
\usepackage[margin=1.0in]{geometry}
\usepackage{graphicx}
\usepackage{amsmath}
\usepackage{amssymb}
\usepackage{bm}
\DeclareMathOperator*{\argmax}{argmax}
\usepackage{physics}
\usepackage[linesnumbered,ruled,vlined]{algorithm2e}
\SetKwInput{KwInput}{Input}                
\SetKwInput{KwOutput}{Output}   
\usepackage[numbers,sort&compress]{natbib}
\usepackage{multirow}
\usepackage{caption} 
\usepackage{enumitem}
\usepackage[pagebackref=true,breaklinks=true,colorlinks,bookmarks=false]{hyperref}

\usepackage{abstract}

\usepackage{titlesec}
\titleformat{\section}{\large\bfseries}{\thesection}{1em}{}
\titleformat{\subsection}{\normalsize\bfseries}{\thesubsection}{1em}{}

\title{\textbf{Exploiting Frequency Spectrum of Adversarial Images for General Robustness}}
\author {
    Chun Yang Tan,\footnote{Graduate School of Science and Engineering, Chiba University (email: chunyangtan@gmail.com)} 
    \ Kazuhiko Kawamoto,$^{\dagger}$
    \ Hiroshi Kera\footnote{Graduate School of Engineering, Chiba University (email: kawa@faculty.chiba-u.jp, kera.hiroshi@gmail.com)} 
}
\date{}

\begin{document}

\maketitle

\begin{abstract}
\textit{
   In recent years, there has been growing concern over the vulnerability of convolutional neural networks (CNNs) to image perturbations. However, achieving general robustness against different types of perturbations remains challenging, in which enhancing robustness to some perturbations (e.g., adversarial perturbations) may degrade others~(e.g., common corruptions). In this paper, we demonstrate that adversarial training with an emphasis on phase components significantly improves model performance on clean, adversarial, and common corruption accuracies. We propose a frequency-based data augmentation method, Adversarial Amplitude Swap, that swaps the amplitude spectrum between clean and adversarial images to generate two novel training images: adversarial amplitude and adversarial phase images. These images act as substitutes for adversarial images and can be implemented in various adversarial training setups. Through extensive experiments, we demonstrate that our method enables the CNNs to gain general robustness against different types of perturbations and results in a uniform performance against all types of common corruptions.   }
\end{abstract}

\section{Introduction}

Despite their state-of-the-art performance~\cite{krizhevsky2012imagenet,lecun1998gradient}, 
convolutional neural networks~(CNNs) have been found to be vulnerable to image perturbations, such as common corruptions~\cite{vasiljevic2016examining,hendrycks2019corruption} and adversarial perturbations~\cite{szegedy2014intriguing,carlini2017towards,carlini2017adversarial,narodytska2017blackbox}. 
These perturbations can lead to a substantial degradation in the performance of CNN classifiers, which can be especially problematic in real-world applications where robustness is critical~\cite{kurakin2018adversarialphysical, papernot2017practicalblackbox,eykholt2018physicalworldattack}.
Data augmentation is often used to improve model robustness against perturbations~\cite{zheng2016improving,hendrycks2020augmix,chen2021apr, rusak2020advnoisettraining}. 
However, existing data augmentation methods rarely improve robustness across all types of perturbations. 
For example, data augmentation with Gaussian noise can enhance robustness against certain corruptions (e.g., \textit{Gaussian noise} and \textit{shot noise}). 
However, this method does not generalize across all corruptions (e.g., \textit{fog} and \textit{contrast})~\cite{geirhos2018generalisation, ford2019advnoise}. 
More importantly, a model which is robust to common corruptions is not guaranteed to be robust against adversarial perturbations~\cite{rusak2020advnoisettraining}. 
This is because common corruptions are average-case perturbations, while adversarial perturbations are the worst-case perturbations. 
To boost the robustness against such worst-case perturbations, adversarial training, which involves training with adversarial images, is often used~\cite{madry2018towards,zhang2019trades}. 
However, training with such images deteriorates model performance on clean images and those perturbed with certain types of common corruptions (e.g., \textit{fog} and \textit{contrast})~\cite{fazwi2018adversarial,shafahi2018inevitable, tsipras2018oddwithacc}. 

To address the general robustness of CNNs across diverse perturbations, recent studies have focused on the more intrinsic properties that lead to the disparities in human and CNN recognition. 
While human recognition~(i.e., generally robust recognition) is strongly biased towards shape information, CNNs tend to bias towards texture information~\cite{geirhos2019texturebias}. 
From the frequency perspective, shape bias can be related to the phase components, which retain most of the high-level semantics in the original images, while texture bias can be related to the amplitude components, which mainly contain low-level statistics~\cite{oppenheim1979phase, oppenheim1981phaseimpor, li2015phasesaliency, leon1982visual}. 
Hence, a generally robust classifier should be able to capture the semantic information from the phase components while being robust against variance in the amplitude~\cite{chen2021apr}. 
Several studies have leveraged this idea and demonstrated that swapping some, if not all, of the amplitude spectrum with other images can encourage CNNs to learn more from the phase components rather than the amplitude~\cite{yang2020fourierda, xu2021domaingeneralization, chen2021apr}. 
\citet{chen2021apr} proposed a frequency-based data augmentation method to swap the amplitude spectrum of an image with that of another, thereby encouraging CNNs to learn more from the phase components instead of the amplitude.
Their method improves the robustness against common corruptions but has little effect to adversarial perturbations. 

In this paper, we show that adversarial training with an emphasis on phase components significantly boosts its performance in various aspects. 
Particularly, it improves the clean, adversarial, and common corruption accuracies and prevents widely known overfitting issues (i.e., catastrophic and robust overfitting). 
Furthermore, we observed that the model trained with our method shows uniform performances over all types of common corruptions, making it more reliable in practical applications.
Our method, Adversarial Amplitude Swap~(AAS), swaps the amplitude spectrum between clean and adversarial images to generate two novel training images, i.e., adversarial amplitude (AA) and adversarial phase (AP) images. 
Our method can be implemented in various adversarial training setups to generate substitutes for the original adversarial images. 
We demonstrate that these images enables CNNs to better extract semantic information from adversarial images, leading to generally robust classifiers.   

Our main contributions can be summarized as follows:
\begin{itemize}[topsep=2pt]
\setlength\itemsep{0pt}
\item We propose a frequency-based data augmentation method, AAS, that generates substitutes for adversarial images to attain general robustness. 
\item We demonstrate that AAS improve model performance in various aspects. In particular, ResNet-50 classifier trained with AAS on CIFAR-10~\cite{krizhevsky2009multilayersfeatures} under PGD adversarial training~\cite{madry2018towards} achieved improvements on clean~(+4.6\%), adversarial~(+6.1\%), and common corruption~(+4.9\%) accuracies. 
\item We demonstrate that AAS can prevent catastrophic and robust overfitting in adversarial training. Furthermore, the model trained with AAS performs uniformly on all types of common corruptions~(only 6.5\% difference between the highest and lowest).  
\end{itemize}

\section{Related Work}

\paragraph{The utility of phase components for robustness.}
One reason CNNs are less robust than humans is the difference in how they process images~\cite{ilyas2019notbugbutfeatures,wang2020highfrequency}. 
Earlier studies have shown that the semantic information required for humans recognition is contained more in the phase spectrum instead of the amplitude. 
However, \cite{chen2021apr} pointed out an unintuitive behavior of CNNs, in which the predictions of CNNs sometimes rely on the amplitude spectrum. 
Several works have exploited the phase components of images to improve robustness~\cite{yang2020fourierda,xu2021domaingeneralization,chen2021apr}. 
\citet{yang2020fourierda} proposed an image translation strategy by replacing or mixing the amplitude spectrum of a source image with that of a random target image in the domain adaptation task. 
\citet{xu2021domaingeneralization} further extended the application to domain generalization task. 
These works have demonstrated that by encouraging CNNs to learn more from the phase components, CNNs can better extract the semantic information of different objects that are robust to domain shifts. 
The most similar to ours, \citet{chen2021apr} proposed a data augmentation strategy to swap the amplitude spectrum of an image with another randomly-picked image. 
Their method improves common corruption robustness, with little effect on the robustness against adversarial perturbations.
In this study, we extended the application of emphasizing the phase components to adversarial training. 
We demonstrate that this simple strategy can significantly improve model performance in various aspects.

\paragraph{Trade-off between high and low-frequency domain.}
A line of work has attempted to explain the robustness of CNNs by exploiting the frequency bandwidth~\cite{huang2021fsdr,mukai2022ood,lin2022freqbias}. 
Particularly, \citet{wang2020highfrequency} suggested that CNNs exploit high-frequency components that are not perceptible to humans to achieve high generalization, leading to the vulnerability of CNNs to high-frequency perturbations. 
Several works attempted to bias CNNs towards the low-frequency components to gain robustness against high-frequency perturbations~\cite{chan2022frequencyrobustness, saikia2021freqbiased}. 
However, this approach does not fully solve the the issue, as perturbations can also be designed by restricting to the low-frequency domain~\cite{guo2020lowfreqperturbation, sharma2019effectivenesslow}. 
The existence of trade-offs between the perturbations in the high and the low-frequency domain makes it more challenging to achieve the goal of general robustness against all types of perturbation by only focusing on the frequency bandwidth~\cite{yin2019fourierperspective, maiya2021frequency,saikia2021freqbiased,fujii2022unsupervised,tsuzuku2019fourierbasis}.
Hence, instead of exploiting the frequency bandwidth, we attempt to exploit the amplitude and phase spectrum for general robustness.

\section{A Closer Look at the Fourier Spectra}\label{sec:fourierspecrum}
Previous studies have shown encouraging outcomes in training robust CNNs by swapping parts of the amplitude spectrum between different images~\cite{xu2021domaingeneralization,yang2020fourierda,chen2021apr}. 
In this section, we aim to delve deeper into these amplitude-swapped images and provide more insights on it. 

\subsection{Phase Spectrum Contains Semantic Information.}
Hereinafter, all the operations (e.g., multiplication and exponentiation) except the discrete Fourier transform (DFT)~\cite{bracewell1986fouriertransform} and its inverse (IDFT) will be applied element-wise.
Let $\bm{x}$ be an $N\times N$ image, $\mathcal{F}$ be the DFT, and $\mathcal{F}^{-1}$ be the IDFT. 
The amplitude spectrum $\mathcal{A}(\bm{x})$ and the phase spectrum $\mathcal{P}(\bm{x})$ of image $\bm{x}$ are represented as follows.
\begin{align}
    \mathcal{A}(\bm{x}) &= \sqrt {\text{Re}[\mathcal{F}(\bm{x})]^2 + \text{Im}[\mathcal{F}(\bm{x})]^2},\label{eq:amp}
    \\
    \mathcal{P}(\bm{x}) &= \arctan \left( \frac{\text{Im}[\mathcal{F}(\bm{x})]}{\text{Re}[\mathcal{F}(\bm{x})]} \right),\label{eq:phase}
\end{align}
where the discrete $\text{Re}[\mathcal{F}(\bm{x})]$ and $\text{Im}[\mathcal{F}(\bm{x})]$ denote the real and imaginary parts of $\mathcal{F}(\bm{x})$, respectively.  
For RGB images, the DFT is performed for each channel to obtain the corresponding amplitude and phase spectrum. 

\begin{figure}[t]
    \centering
    \includegraphics[width=0.35\textwidth]{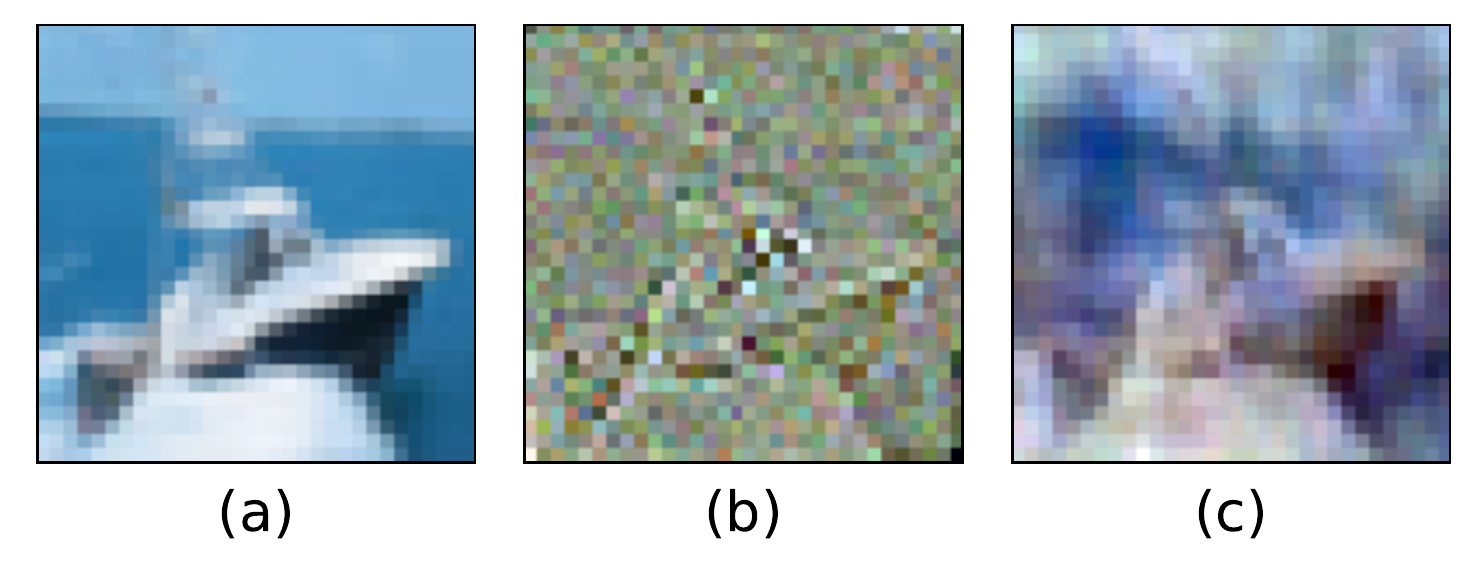}
    \caption{Examples of images with different amplitude components. (a) Clean image, (b) $\bm{x}_\text{phase}$ image consists of only the phase spectrum of clean image, (c) $\bm{x}_\text{swap}$ image consists of the phase spectrum of clean image and amplitude spectrum of a randomly-picked image. }
    \label{fig:phaseimages}
\end{figure}

We reconstructed several images by altering the amplitude spectrum while retaining the phase: 
(i) clean images $\bm{x}$ without any modification, 
(ii) phase images $\bm{x}_\text{phase}$, which contain only the phase spectrum of $\bm{x}$ with amplitude spectrum set to a constant value of 1, and
(iii) swap images $\bm{x}_\text{swap}$, which contain the phase spectrum of $\bm{x}$ and the amplitude spectrum of another image $\bm{x}_m$ randomly picked from the same batch of training images. Formally, phase images  $\bm{x}_\text{phase}$ and swap images $\bm{x}_\text{swap}$ with newly-adopted spectra can be represented as follows:
\begin{align}
    \bm{x_\text{phase}} &= \mathcal{F}^{-1} \left(\bm{1} \cdot e^{i\mathcal{P}(\bm{x})}\right), \\
    \bm{x_\text{swap}} &= \mathcal{F}^{-1} \left(\mathcal{A}\cdot (\bm{x}_m) \ e^{i\mathcal{P}(\bm{x})}\right).
\end{align} 
Some examples of the generated images are shown in Figure~\ref{fig:phaseimages}. As observed, all the images still retain the semantic information needed to be classified as a \textit{ship}. However, from a human's perspective, both the images $\bm{x}_\text{phase}$ and $\bm{x}_\text{swap}$ are heavily perturbed and are challenging to recognize. More examples of swap images are shown in the supplementary material. 

Interestingly, CNNs can learn useful features to classify \textit{clean} images from such barely recognizable images despite the large discrepancy between them. 
To show this, we trained three ResNet-50 models using each image independently. Table~\ref{table:reconstructimage} shows the results. 
Notably, both the models trained with phase images ($\bm{x}_\text{phase}$) and swap images ($\bm{x}_\text{swap}$) performed reasonably well on clean test images. 
Although $\bm{x}_\text{phase}$ does not contain any amplitude components of the clean images, the model could still pick up necessary information from the phase spectrum and performed well on clean test images, with just a 6.8\% drop in accuracy. 
This outcome indicates that the phase spectrum contains most of the semantic information required for image classification. 
Another interesting outcome is that despite having totally unrelated amplitude spectrum, the model trained with $\bm{x}_\text{swap}$ still managed to pick up the necessary information and generalize well on clean test images, with just a 1.4\% drop in accuracy. 
Moreover, this model showed a comparable common corruption robustness (+0.7\%) with the standard model trained on clean training images. 
Note that these swap images $\bm{x}_\text{swap}$ are also known as \mbox{APR-P} images in \cite{chen2021apr}. 
Their method \mbox{APR-P} involves training with both the clean images and the swap images to encourage the model to learn more from the phase spectrum, rather than the amplitude. 
As observed, the model trained with a combination of clean and swap images~(\mbox{APR-P}) achieved higher common corruption robustness (+1.8\%) with comparable clean accuracy compared to the standard model trained with only clean images. 
Here, we emphasize that despite having unrelated amplitude components, swap images do contain enough information required for generalization. 

\setlength{\tabcolsep}{4pt}
\begin{table}[tb]
    \centering
    \begin{tabular}{cccc}
        \hline
        \multicolumn{1}{c|}{Training}& \multicolumn{3}{c}{Accuracy (\%)} \\
        \multicolumn{1}{c|}{Data}&Clean&PGD-$l_\infty$&Corruption\\
        \hline
        \multicolumn{1}{c|}{$\bm{x}$} &   
            94.1&0.0&85.2\\
        \multicolumn{1}{c|}{$\bm{x_\text{phase}}$} &
            87.2&15.2&81.5\\
        \multicolumn{1}{c|}{$\bm{x_\text{swap}}$} &
            92.7&0.2&86.5\\
        \multicolumn{1}{c|}{APR-P~\cite{chen2021apr}} &
            93.8&0.6&87.4\\
        \hline
    \end{tabular}
    \caption{The classification accuracy of ResNet-50 classifiers trained with images with the same phase spectrum but different amplitude spectrum. Evaluation was done on clean, PGD-perturbed, and commonly-corrupted images.}
    \label{table:reconstructimage}
\end{table}

\subsection{Phase-related Information in Swap Images} \label{sec:capture-phase}

To further understand the behavior of CNNs in capturing generalizable information from swap images $\bm{x}_\text{swap}$, we conducted another experiment. 
We demonstrate that the phase spectrum contains generalizable information related to the labels. 
In contrast, CNN models struggle to capture sufficient information related to the labels for generalization from the amplitude spectrum. 

Consider the swapping process between clean training images $\bm{x}_p$ with labels $y_p$, and images randomly drawn from the same training batch $\bm{x}_q$ with labels $y_q$ to generate swap images $\bm{x}_\text{swap}$. 
These swap images adopt frequency components from two different images, i.e., the phase spectrum from $\bm{x}_p$ and the amplitude spectrum from $\bm{x}_q$. To demonstrate whether CNNs can capture generalizable information from each of the frequency components, we trained three ResNet-50 models by using (i) $y_p$, (ii) $y_q$ and (iii) both $y_p$ and $y_q$ as targets. 
The results are shown in Table~\ref{table:difflabels}. 
First, the training with (i) $y_p$ as targets corresponds to the training with swap images ($\bm{x}_\text{swap}$) shown in Table~\ref{table:reconstructimage}. 
The model managed to capture generalizable information related to the labels associated with the phase components, leading to an accurate and robust classifier. 
The training with (ii) $y_q$ as targets, on the other hand, suffered from a drastic drop in performance on clean test images (51.7\%). 
The model struggled to capture generalizable features from swap images corresponding to the amplitude spectrum. 
Interestingly, without any additional restriction, the model trained with (iii) both $y_p$ and $y_q$ as targets managed to learn and generalize well on clean and corrupted images, showing comparable performance with the model trained with (i) $y_p$ as targets. 
Although the model was allowed to capture the information corresponding to either amplitude or phase spectrum, the model still managed to pick up generalizable features, which is believed to be mainly contained in the phase spectrum. 
Despite having unrelated amplitude spectrum, CNNs can efficiently capture generalizable information from the phase spectrum of these swap images. 
These outcomes consolidated the idea of training a robust classifier by altering the amplitude spectrum of the training images. 

\setlength{\tabcolsep}{4pt}
\begin{table}[tb]
    \centering
    \begin{tabular}{ccccc}
        \hline
        \multicolumn{1}{c|}{Training}&\multicolumn{1}{c|}{\multirow{2}{*}{Labels}}& \multicolumn{3}{c}{Accuracy (\%)} \\
        \multicolumn{1}{c|}{Data}&\multicolumn{1}{c|}{}&Clean&PGD-$l_\infty$&Corruption\\
        \hline
        \multicolumn{1}{c|}{\multirow{3}{*}{$\bm{x}_\text{swap}$}}&\multicolumn{1}{c|}{$y_p$} &
            \textbf{92.7}&0.2&\textbf{86.5}\\
        \multicolumn{1}{c|}{}&\multicolumn{1}{c|}{$y_q$} &
            51.7&0.0&45.5\\
        \multicolumn{1}{c|}{}&\multicolumn{1}{c|}{$y_p, y_q$} &
            92.3&\textbf{1.5}&84.9\\
        \hline
    \end{tabular}
    \caption{The classification accuracy of ResNet-50 classifiers trained with swap images with different labels as targets. Evaluation was done on clean, PGD-perturbed, and commonly-corrupted images. The best results are indicated in bold.}
    \label{table:difflabels}
\end{table}

\section{Method}

The outcomes in Section~\ref{sec:fourierspecrum} suggest that CNNs can efficiently learn from swap images, which contain randomly-picked amplitude spectrum. 
In this section, we aim to explore
\textit{swap images with adversarially-perturbed amplitude spectrum}.
Here, we propose Adversarial Amplitude Swap~(AAS), a frequency-based data augmentation method that constructs swap images from clean and adversarial images. 
The method generates substitutes for adversarial images and can be implemented in any adversarial training. 

\subsection{Revisiting Adversarial Training}

\paragraph{Adversarial images.} 
Let $\mathcal{X}$ be the image domain, and let $\mathcal{Y}=\{1, 2, \ldots, K\}$. Let $F(\bm{x}) = \argmax_{i\in\mathcal{Y}} f_i(\bm{x})$ be a $K$-class CNN classifier, where $f_i:\mathcal{X}\to\mathbb{R}$ denotes the $i$-th logit of the CNN. Given a distance function $d:\mathcal{X}\times \mathcal{X}\to \mathbb{R}_{\ge 0}$ and a budget $\epsilon > 0$, an adversarial image $\bm{x}_{\text{adv}}\in\mathcal{X}$ of $\bm{x}\in\mathcal{X}$ with the label $y\in\mathcal{Y}$ is an image such that $F(\bm{x}_{\text{adv}}) \ne y$ and $d(\bm{x}, \bm{x}_{\text{adv}}) \le \epsilon$.
Gradient-based non-targeted methods to generate adversarial examples typically solve the following problem approximately based on gradient ascent with some projections. 
\begin{align}\label{eq:grad-based-adv}
    \max_{\bm{x}'\in\mathcal{X}}\ \ell(f(\bm{x}'), y),\quad\text{s.t.}\  d(\bm{x},\bm{x}') \le \epsilon,
\end{align}
where $f(\bm{x}) = (f_1(\bm{x}), \ldots, f_K(\bm{x}))^{\top}$, and $\ell(\,\cdot\,,\,\cdot\,)$ denotes a loss function (e.g., cross-entropy loss). 

\paragraph{Adversarial training.}
Adversarial training~\cite{szegedy2014intriguing} is a defense method in which a classifier is trained to classify adversarial images correctly. 
Standard adversarial training aims to solve the following min-max optimization problem. 
\begin{align}\label{eq:std-adv-training}
    \min_{f} \mathbb{E} \left\{\max_{\bm{x}'\in\mathcal{X}}\ \ell(f(\bm{x}'), y)\right\}.
\end{align}
Another variant of adversarial training approach, TRADES, has been proposed by~\citet{zhang2019trades}. TRADES combines the idea of adversarial training and regularization term, to encourage the model to have a better trade-off between clean accuracy and adversarial robustness. TRADES aims to solve the following optimization problem.
\begin{align}\label{eq:trades-adv-training}
    \min_{f} \mathbb{E} \left\{\ell(f(\bm{x}), y)  +  \beta\cdot\max_{\bm{x}'\in\mathcal{X}}\ \ell(f(\bm{x}), f(\bm{x}'))\right\},
\end{align}
where $\beta$ serves as a hyper-parameter to control the trade-off between clean accuracy and adversarial robustness. 
In either of the case in Eq.~\eqref{eq:std-adv-training} and Eq.~\eqref{eq:trades-adv-training}, the fast gradient sign method~(FGSM)~\cite{goodfellow2015explaining} and the projected gradient descent~(PGD)~\cite{madry2018towards} are two commonly-used algorithms for solving the inner maximization problem.

\paragraph{Overfitting in adversarial training.}

FGSM adversarial training is a computationally inexpensive approach. However, it is known that when $\epsilon$ is too large, the classifier starts performing unreasonably well for FGSM adversarial images and fails to classify other adversarial images (e.g., those generated by PGD). This is known as catastrophic overfitting~\cite{tramer2018ensembleattdef,kim2020understandingco, kang2021understandingco,wong2020fast,andriushchenko2020improvefast}.
Another form of overfitting that may occur in adversarial training is robust overfitting~\cite{rice2020overfitting, tack2021consistency}, where the classification accuracy of adversarial images starts degrading substantially at some point in an adversarial training with the PGD~\cite{madry2018towards}. 
The development of methods to circumvent these forms of overfitting and train a strongly defended classifier remains a challenging task. 

\begin{figure}[t]
    \centering
    \includegraphics[width=0.45\textwidth]{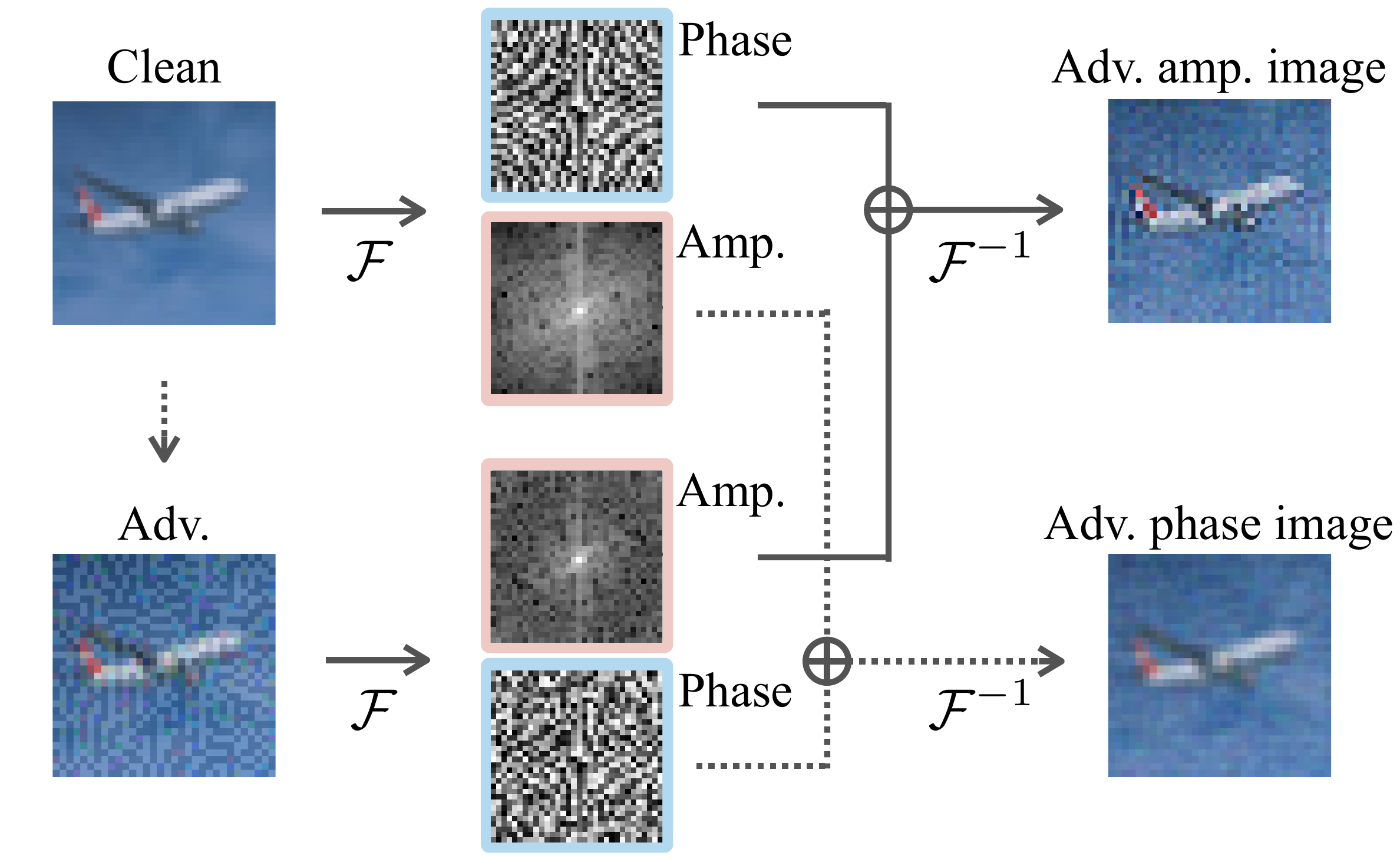}
\caption{The pipeline of Adversarial Amplitude Swap~(AAS). The maps $\mathcal{F}$ and $\mathcal{F}^{-1}$ denote the discrete Fourier transform and its inverse. The amplitude spectra of a clean image and its adversarial image are swapped to generate adversarial amplitude~(AA) and adversarial phase~(AP) images. 
}\label{fig:teaser}
\end{figure}

\subsection{Adversarial Amplitude Swap}

Adversarial training improves adversarial robustness but substantially deteriorates the model performance on clean images~\cite{fazwi2018adversarial, shafahi2018inevitable, tsipras2018oddwithacc} and common corruptions~\cite{laugros2019adversarial, laugros2020addressing}.
To better address the general robustness of CNNs against different types of perturbations, we propose Adversarial Amplitude Swap~(AAS), a frequency-based data augmentation method to encourage CNNs to learn more from the phase components in adversarial training.
Given a clean and its adversarial images, this method swaps the amplitude spectrum of the former with that of the latter to generate two augmented images: an adversarial amplitude (AA) image, which has the amplitude spectrum of the adversarial image and the phase spectrum of the clean image, and an adversarial phase (AP) image, which is the opposite~(Figure~\ref{fig:teaser}). 

Formally, the process of AAS is performed as follows.
First,  given a clean image $\bm{x}$, an adversarial image $\bm{x}_{\text{adv}}$ is generated. Then, the DFT is applied to the two images to obtain the \mbox{amplitude-phase} decompositions, $(\mathcal{A}(\bm{x}), \mathcal{P}(\bm{x}))$ and $(\mathcal{A}(\bm{x}_{\text{adv}}), \mathcal{P}(\bm{x}_{\text{adv}}))$.
AA and AP images are then constructed by the IDFT of $(\mathcal{A}(\bm{x}), \mathcal{P}(\bm{x}_{\text{adv}}))$ and $(\mathcal{A}(\bm{x}_{\text{adv}}), \mathcal{P}(\bm{x}))$, respectively; namely, 
\begin{align}
    \bm{x}_{\text{AA}} &= \mathcal{F}^{-1}\qty(\mathcal{A}(\bm{x}_{\text{adv}})\cdot e^{i \mathcal{P}(\bm{x})}), \label{eq:aa}\\
    \bm{x}_{\text{AP}} &= \mathcal{F}^{-1}\qty(\mathcal{A}(\bm{x})\cdot e^{i\mathcal{P}(\bm{x}_{\text{adv}})}).\label{eq:ap}
\end{align}
Pseudo-code for this process is provided in Algorithm~\ref{alg:AAS}. 
Note that the adversarial images are generated by solving the inner maximization problem in Eq.~\eqref{eq:std-adv-training} for standard adversarial training and in Eq.~\eqref{eq:trades-adv-training} for TRADES adversarial training. 
Hence, adversarial images contain stochastic frequency spectrum that changes at each training step. 

Our method focuses on training CNN models with AA images, which consist of the static phase spectrum derived from the clean images, and the stochastic amplitude spectrum derived from the adversarial images. 
The stochastic amplitude spectrum in AA images, which changes at each training step, enables the models to better extract semantic information from the static phase spectrum.
On the other hand, AP images consist of the stochastic phase spectrum derived from the adversarial images and the static amplitude spectrum derived from the clean images. 
The training with AP images is somewhat similar to training with original adversarial images, except that AP images contain only the stochastic phase spectrum. 
Note that AAS can be implemented in any kind of adversarial training as both the AA and AP images act as substitutes for the adversarial images. 

In Section~\ref{sec:experiment-classification}, we demonstrated that the training with either AA or AP images leads to an overall improvement in various aspects compared to the standard adversarial training setup. 
Particularly, AA images, which contain static phase spectrum with stochastic amplitude spectrum, enable CNN models to better capture semantic information from the phase components while learning to resist the adversarial features in the amplitude spectrum. 
This eventually leads to classifiers more general robustness.
In particular, the model trained with AA images not only performs better on clean and adversarial images but also performs uniformly against different types of common corruptions(Section~\ref{sec:experiment-generalrobustness}). 
The uniform performances in all corruption types indicate that the model is more likely to perform similarly on unseen corruptions, which is more favorable in real-world scenarios. 
In Section~\ref{sec:experiment-overfitting}, we demonstrate that our method can also help stabilize adversarial training to circumvent catastrophic and robust overfitting. 

\begin{algorithm}[t]
\DontPrintSemicolon
  \KwInput{$\bm{x}$: clean image}
  \KwOutput{ $\bm{x}_{\text{AA}}$: adversarial amplitude image, \\ \hspace{39pt} $\bm{x}_{\text{AP}}$: adversarial phase image}
  $\bm{x}_{\text{adv}} \leftarrow \text{AdversarialAttack}(\bm{x})$. \\
  $\mathcal{A}(\bm{x}), \mathcal{P}(\bm{x}) \leftarrow \text{DFT}(\bm{x})$ \\ 
  $\mathcal{A}(\bm{x}_{\text{adv}}), \mathcal{P}(\bm{x}_{\text{adv}}) \leftarrow \text{DFT}(\bm{x}_{\text{adv}})$ \\ 
  $\bm{x}_{\text{AA}} \leftarrow \text{IDFT}(\mathcal{A}(\bm{x}_\text{adv}), \mathcal{P}(\bm{x}))$ \\
  $\bm{x}_{\text{AP}} \leftarrow \text{IDFT}(\mathcal{A}(\bm{x}), \mathcal{P}(\bm{x}_{\text{adv}}))$ 
\caption{Adversarial Amplitude Swap (AAS)}\label{alg:AAS}
\end{algorithm}

\section{Expeiments}

\subsection{CIFAR-10 Image Classification} \label{sec:experiment-classification}
\paragraph{Training setup.}
We conducted several experiments on CIFAR-10 dataset~\cite{krizhevsky2009multilayersfeatures}. 
For every training, the learning rate was set to 0.01 with a decay of 0.1 at the \mbox{100-th} and \mbox{150-th} epochs. The classifiers were optimized with stochastic gradient descent using a momentum of 0.9, and a weight decay of $5\times 10^{-4}$. Two conventional data augmentation methods, including random crop and random horizontal flip, were implemented in the training of all models unless stated otherwise. 
We evaluated the effectiveness of our method in enhancing adversarial training on several different adversarial training setups. 
We also conducted experiments on different dataset~(CIFAR-100~\cite{krizhevsky2009multilayersfeatures}) and different network architectures~(WideResNet~\cite{zagoruyko2016wrn} and DenseNet~\cite{huang2017densenet}) and observed the same trends. Due to page limitations, we send the results to the supplementary material. 

\paragraph{Evaluation.} 
We evaluate model performance on the CIFAR-10 test dataset. 
To evaluate model robustness against adversarial perturbations, 
we used Pytorch~\cite{paszke2019pytorch} library, Foolbox~\cite{rauber2020foolbox}, to generate 
FGSM perturbation with \mbox{$\epsilon=\frac{8}{255}$}, \mbox{PGD-$l_\infty$} perturbation with \mbox{$\epsilon=\frac{8}{255}$}, the step size \mbox{$\alpha=\frac{2}{255}$}, and the number of iterations \mbox{$n=20$}. 
To evaluate model robustness against common corruptions, we evaluated methods on the CIFAR-10-C dataset~\cite{hendrycks2019corruption}.
We demonstrate the average accuracy of models on a all types of common corruptions (Corr), which include \textit{Gaussian noise, shot noise, impulse noise, defocus blur, frosted, glass blur, motion blur, zoom blur, snow, frost, fog, brightness, contrast, elastic, pixelate, JPEG compression}, each appearing at five severity levels or intensities.

\setlength{\tabcolsep}{8pt}
\begin{table}[tb]
    \centering
    \begin{tabular}{ccccc}
        \hline
        \multicolumn{1}{c|}{}& \multicolumn{4}{c}{Accuracy (\%)} \\
        \multicolumn{1}{c|}{}&Clean&FGSM&PGD&Corr\\
        \hline
        \hline
        \multicolumn{1}{c|}{Adv} &   
            82.8&60.0&44.4&78.9\\
        \multicolumn{1}{c|}{AA} &
            \textbf{89.1}&\textbf{74.6}&\textbf{50.1}&\textbf{85.7}\\
        \multicolumn{1}{c|}{AP} &
            87.4&72.0&46.3&83.8\\
        \hline
        \multicolumn{1}{c|}{C\&Adv} &   
            84.7&61.4&54.3&80.8\\
        \multicolumn{1}{c|}{C\&AA} &
            \textbf{89.3}&\textbf{74.3}&\textbf{60.4}&\textbf{85.7}\\
        \multicolumn{1}{c|}{C\&AP} &
            86.9&70.5&56.8&83.3\\
        \hline
    \end{tabular}
        \caption{The classification accuracy of ResNet-50 classifiers trained with PGD adversarial training~\cite{madry2018towards}. The best results are indicated in bold.}
    \label{table:resnet50pgd}
\end{table}

\paragraph{Results on standard adversarial training.}

First, we trained ResNet-50~\cite{he2016resnet} models with standard adversarial training~\cite{madry2018towards}, which involves solving the inner maximization problem in Eq.~\eqref{eq:std-adv-training} by using the PGD algorithm with $l_\infty$ norm with \mbox{$\epsilon=\frac{8}{255}$}, the step size $\alpha=\frac{2}{255}$ and the number of iterations $n=10$. 
The results are shown in Table~\ref{table:resnet50pgd}. 
Compared to the baseline model trained with adversarial images~(Adv), both the training with adversarial amplitude~(AA) images and the training with adversarial phase~(AP) images achieved overall improvements in all aspects, including clean, adversarial, and common corruption accuracies. 
Notably, training with AA images with the purpose of encouraging the model to learn from the static phase spectrum achieved the best performance in all aspects (+6.3\% on clean, +14.6\% on FGSM, +5.7\% on PGD, and +6.8\% on common corruption accuracies). 
Next, we focus on the training with a mixture of clean and adversarial images. 
The training with our method, C\&AA and C\&AP, consistently outperformed the baseline model (C\&Adv).
Particularly, the training includes AA images (C\&AA) outperformed the baseline model in all aspects (+4.6\% on clean, +12.9\% on FGSM, +6.1\% on PGD, and +4.9\% on common corruption accuracies). 

\setlength{\tabcolsep}{3pt}
\begin{table}[t]
    \centering
    \begin{tabular}{ccccc}
        \hline
        \multicolumn{1}{c|}{}& \multicolumn{4}{c}{Accuracy (\%)} \\
        \multicolumn{1}{c|}{}&Clean&FGSM&PGD&Corr\\
        \hline
        \hline
        \multicolumn{1}{c|}{TRADES ($\beta=1$)} &   
            84.3&54.8&43.5&80.5\\
        \multicolumn{1}{c|}{TRADES ($\beta=1$)+AA} &
            \textbf{89.9}&\textbf{70.5}&\textbf{48.1}&\textbf{86.4}\\
        \multicolumn{1}{c|}{TRADES ($\beta=1$)+AP} &
            88.2&68.5&44.4&84.9\\
        \hline
        \multicolumn{1}{c|}{TRADES ($\beta=3$)} &   
            82.5&54.3&43.1&79.1\\
        \multicolumn{1}{c|}{TRADES ($\beta=3$)+AA} &
            \textbf{88.0}&\textbf{67.4}&\textbf{51.8}&\textbf{84.6}\\
        \multicolumn{1}{c|}{TRADES ($\beta=3$)+AP} &
            85.8&66.0&50.2&82.8\\
        \hline
        \multicolumn{1}{c|}{TRADES ($\beta=6$)} &   
            80.7&54.1&44.5&77.2\\
        \multicolumn{1}{c|}{TRADES ($\beta=6$)+AA} &
            \textbf{87.3}&62.9&46.4&\textbf{83.9}\\
        \multicolumn{1}{c|}{TRADES ($\beta=6$)+AP} &
            84.9&\textbf{63.0}&\textbf{47.4}&81.4\\
        \hline
    \end{tabular}
        \caption{The classification accuracy of ResNet-18 trained with TRADES~\cite{zhang2019trades} under different $\beta$. PGD attack are used to generate adversarial images during training. The best results under the same $\beta$ for each aspect are indicated in bold.}
    \label{table:resnet18pgd}
\end{table}

\paragraph{Results on TRADES.}
To demonstrate that AAS can also be implemented on different adversarial training setups, we conducted experiments by training Resnet-18 models with TRADES, which generates adversarial images by solving the inner maximization problem in Eq.~\eqref{eq:trades-adv-training}. 
We trained models with three different values of the hyper-parameter~$\beta$. 
The results are shown in Table~\ref{table:resnet18pgd}. 
We observed the same outcomes as that of standard adversarial training (Table~\ref{table:resnet18pgd}). 
All the models trained with AAS consistently outperformed the baseline models in all aspects under the same training setup. 
In particular, the model trained with AA images achieved the best performance in all aspects (+5.5\% on clean, +13.1\% on FGSM, +8.7\% on PGD, and +5.5\% on common corruption accuracies) compared to the baseline model. 

On the whole, AAS enables CNN models to gain more adversarial robustness while retaining clean and common corruption accuracies. 
In particular, AA images enable the models to better extract semantic information from the phase spectrum and hence lead to more generally robust classifiers. 
In addition to that, AA images also offer a more uniform performance on all common corruption types, which will be discussed in Section~\ref{sec:experiment-generalrobustness}. 

\paragraph{Adversarial features of AA and AP images.}
We demonstrated the effectiveness of AA and AP images in enhancing model performance in adversarial training and training classifiers with more general robustness against different types of perturbations. 
Notably, despite the ability to improve adversarial robustness, AA images did not serve as a strong adversarial attack compared to original adversarial images, as shown in Table~\ref{table:errorrate}. 
We tested several classifiers trained with clean images and showed the classification errors of the classifiers on clean, adversarial (Adv), adversarial amplitude (AA), and adversarial phase (AP) images. 
We noticed that the original adversarial images served as the strongest attack. 
Both AA and AP images have a lower fooling rate on the classifiers because they contain only either one of the frequency spectra of adversarial images. 
Despite the weaker fooling ability, both AA and AP images enabled the models to gain improvements in all aspects, including clean, adversarial, and common corruption accuracies. 
The utility of weaker attacks in adversarial training has already been shown in several studies.
\citet{kireev2022advoncorruptions} showed that adversarial training with weaker attacks can serve as a method to moderately enhance the robustness against common corruptions.
\citet{zhang2020fat} showed that the adversarial training with adaptive strength of attacks has better clean accuracy than the original one.
Unlike these studies, which only focused on either common corruptions or adversarial robustness, our study addressed the general robustness, in which the model trained with our method showed improvements in clean, adversarial, and common corruption accuracies. 

\setlength{\tabcolsep}{4pt}
\begin{table}[tb]
    \centering
    \begin{tabular}{cccccc}
        \hline
        \multicolumn{1}{c|}{}&\multicolumn{1}{c|}{\multirow{2}{*}{Networks}}& \multicolumn{4}{c}{Error Rates (\%)} \\
        \multicolumn{1}{c|}{}&\multicolumn{1}{c|}{}&Clean&Adv&AA&AP\\
        \hline
        \multicolumn{1}{c|}{\multirow{3}{*}{FGSM}}&\multicolumn{1}{c|}{ResNet-18} &   
            5.2&34.4&22.6&27.0\\
        \multicolumn{1}{c|}{}&\multicolumn{1}{c|}{WideResNet} &
            6.0&33.2&21.5&26.3\\
        \multicolumn{1}{c|}{}&\multicolumn{1}{c|}{DenseNet} &
            6.6&42.7&29.6&34.8\\
        \hline
        \multicolumn{1}{c|}{\multirow{3}{*}{PGD-$l_\infty$}}&\multicolumn{1}{c|}{ResNet-18} &   
            5.2&99.9&37.8&76.4\\
        \multicolumn{1}{c|}{}&\multicolumn{1}{c|}{WideResNet} &
            6.0&99.8&36.0&76.5\\
        \multicolumn{1}{c|}{}&\multicolumn{1}{c|}{DenseNet} &
            6.6&100.0&50.4&87.4\\
    \end{tabular}
        \caption{The classification error of classifiers trained with clean images on several types of adversarial images.}
    \label{table:errorrate}
\end{table}

\setlength{\tabcolsep}{0.7pt}
\begin{table*}[tb]
    \centering
    \begin{tabular}{ccccccccccccccccccc}
        \hline
        \multicolumn{1}{c|}{}&
        \multicolumn{1}{c|}{Clean}&
        \multicolumn{1}{c|}{PGD}&
        \multicolumn{1}{c|}{Corr}&
        \multicolumn{3}{c|}{Noise}&
        \multicolumn{4}{c|}{Blur}&
        \multicolumn{4}{c|}{Weather}&
        \multicolumn{4}{c}{Digital}         \\
        \multicolumn{1}{c|}{}&
        \multicolumn{1}{c|}{Acc}&
        \multicolumn{1}{c|}{Acc}&
        \multicolumn{1}{c|}{Mean}&
        \multicolumn{1}{c}{Gauss}&
        \multicolumn{1}{c}{Shot}&
        \multicolumn{1}{c|}{Imp}&
        \multicolumn{1}{c}{Def}&
        \multicolumn{1}{c}{Glass}&
        \multicolumn{1}{c}{Mot}&
        \multicolumn{1}{c|}{Zoom}&
        \multicolumn{1}{c}{Snow}&
        \multicolumn{1}{c}{Frost}&
        \multicolumn{1}{c}{Fog}&
        \multicolumn{1}{c|}{Bright}&
        \multicolumn{1}{c}{Contr}&
        \multicolumn{1}{c}{Elast}&
        \multicolumn{1}{c}{Pxl}&
        \multicolumn{1}{c}{JPEG}        \\
        \hline
        \hline
        \multicolumn{1}{c|}{Std} &   
            \multicolumn{1}{c|}{\textbf{94.1}}&\multicolumn{1}{c|}{0.0}&\multicolumn{1}{c|}{85.2}&\multicolumn{1}{c}{75.5}&\multicolumn{1}{c}{80.4}&\multicolumn{1}{c|}{76.0}&\multicolumn{1}{c}{92.2}&\multicolumn{1}{c}{70.6}&\multicolumn{1}{c}{89.3}&\multicolumn{1}{c|}{90.9}&\multicolumn{1}{c}{86.0}&\multicolumn{1}{c}{86.7}&\multicolumn{1}{c}{91.6}&\multicolumn{1}{c|}{93.3}&\multicolumn{1}{c}{92.2}&\multicolumn{1}{c}{86.3}&\multicolumn{1}{c}{88.3}&\multicolumn{1}{c}{79.3}\\
        \multicolumn{1}{c|}{APR} &
            \multicolumn{1}{c|}{93.9}&\multicolumn{1}{c|}{1.7}&\multicolumn{1}{c|}{\textbf{90.0}}&\multicolumn{1}{c}{87.4}&\multicolumn{1}{c}{\textbf{88.8}}&\multicolumn{1}{c|}{\textbf{88.6}}&\multicolumn{1}{c}{\textbf{92.8}}&\multicolumn{1}{c}{\textbf{81.5}}&\multicolumn{1}{c}{\textbf{91.5}}&\multicolumn{1}{c|}{\textbf{92.1}}&\multicolumn{1}{c}{\textbf{90.7}}&\multicolumn{1}{c}{\textbf{91.6}}&\multicolumn{1}{c}{\textbf{93.0}}&\multicolumn{1}{c|}{\textbf{93.5}}&\multicolumn{1}{c}{\textbf{93.4}}&\multicolumn{1}{c}{\textbf{88.8}}&\multicolumn{1}{c}{\textbf{91.3}}&\multicolumn{1}{c}{84.5}\\
        \multicolumn{1}{c|}{Adv} &
            \multicolumn{1}{c|}{82.8}&\multicolumn{1}{c|}{44.4}&\multicolumn{1}{c|}{78.9}&\multicolumn{1}{c}{81.9}&\multicolumn{1}{c}{82.1}&\multicolumn{1}{c|}{78.8}&\multicolumn{1}{c}{80.0}&\multicolumn{1}{c}{77.6}&\multicolumn{1}{c}{77.7}&\multicolumn{1}{c|}{79.9}&\multicolumn{1}{c}{79.2}&\multicolumn{1}{c}{80.6}&\multicolumn{1}{c}{71.0}&\multicolumn{1}{c|}{82.7}&\multicolumn{1}{c}{72.6}&\multicolumn{1}{c}{78.1}&\multicolumn{1}{c}{81.2}&\multicolumn{1}{c}{80.8}\\
        \multicolumn{1}{c|}{AA} &   
            \multicolumn{1}{c|}{89.1}&\multicolumn{1}{c|}{\textbf{50.1}}&\multicolumn{1}{c|}{85.7}&\multicolumn{1}{c}{\textbf{87.6}}&\multicolumn{1}{c}{87.9}&\multicolumn{1}{c|}{82.9}&\multicolumn{1}{c}{86.9}&\multicolumn{1}{c}{81.4}&\multicolumn{1}{c}{85.2}&\multicolumn{1}{c|}{87.0}&\multicolumn{1}{c}{85.6}&\multicolumn{1}{c}{87.4}&\multicolumn{1}{c}{81.7}&\multicolumn{1}{c|}{89.0}&\multicolumn{1}{c}{84.4}&\multicolumn{1}{c}{84.5}&\multicolumn{1}{c}{87.2}&\multicolumn{1}{c}{\textbf{86.6}}\\
        \multicolumn{1}{c|}{AP} &
            \multicolumn{1}{c|}{87.4}&\multicolumn{1}{c|}{46.3}&\multicolumn{1}{c|}{83.8}&\multicolumn{1}{c}{86.2}&\multicolumn{1}{c}{86.5}&\multicolumn{1}{c|}{82.0}&\multicolumn{1}{c}{85.2}&\multicolumn{1}{c}{80.8}&\multicolumn{1}{c}{82.6}&\multicolumn{1}{c|}{85.1}&\multicolumn{1}{c}{83.8}&\multicolumn{1}{c}{85.4}&\multicolumn{1}{c}{79.0}&\multicolumn{1}{c|}{87.2}&\multicolumn{1}{c}{80.5}&\multicolumn{1}{c}{82.6}&\multicolumn{1}{c}{85.8}&\multicolumn{1}{c}{84.9}\\
        \hline
    \end{tabular}
        \caption{The classification accuracy of ResNet-50 classifiers trained with different training images. Evaluation was done on clean images, PGD-perturbed images, and images corrupted by 15 types of common corruptions. Corr denotes the average accuracy across all 15 types of corruption. All values are percentages. The best results are indicated in bold.}
    \label{table:uniform-robustness}
\end{table*}

\subsection{General and Uniform Robustness} \label{sec:experiment-generalrobustness}

To demonstrate that our method, AAS, leads to classifiers with general robustness, we delve deeper into the details of model performance on different types of common corruptions. Table~\ref{table:uniform-robustness} shows the model performances on clean images, PGD images, and 15 types of common corruption images. 
We compared the our method with several baselines, i.e., standard training, APR~\cite{chen2021apr}, and standard PGD adversarial training~\cite{madry2018towards}. 
Both the models with standard training~(Std) and APR training have comparable (the highest) accuracy on clean images. 
However, the high clean accuracy is attributed to the fact that adversarial images were not used in the training. 
This resulted in total vulnerability to PGD perturbations~($<$2\%). 
The model trained with adversarial training, on the other hand, has reasonable robustness against PGD perturbations, but with a drastic drop in performance on clean (-11.3\%) and common corruption~(-6.3\% on average) accuracies. 
Both the models trained with AAS (AA and AP) showed cosistent improvements against all types of perturbations, including clean, PGD and common corruptions compared to the model trained with standard adversarial training. 

Next, we focus on the robustness of models on each type of common corruption. 
The model with standard training performed well on some corruption types (e.g., 93.3\% on \textit{brightness} and 92.2\% on \textit{contrast}). 
However, the performances on other corruptions was less satisfied (e.g., 75.5\% on \textit{Gaussian noise} and 70.6\% on \textit{Glass Blur}), with the gap between the highest and the lowest being 22.7\%. 
APR did improve the overall performance on all corruption types, but there is still a 12\% difference between the highest~(93.5\% on \textit{brightness}) and the lowest~(81.5\% on \textit{glass blur}) accuracies. 
Note that model trained with APR remained totally vulnerable towards PGD perturbations~(1.7\%). 
For standard PGD adversarial training, the gap between the highest~(82.1\% on \textit{shot noise}) and the lowest~(71\% on \textit{fog}) is 11.7\%. 
The training with our method, especially AA images, performed uniformly across all types of corruptions, with just a 6.5\% difference between the highest~(87.9\% on \textit{shot noise}) and the lowest~(81.4\% on \textit{glass blur}) accuracies. 
Taking into consideration that the 15 types of the commom corruptions used in evaluation might not be all the cases in real world, the model with a uniform performance across these 15 types of commom corruptions is more likely to perform similarly on unseen corruptions. 

\begin{figure}[t]
    \centering
    \includegraphics[width=0.45\textwidth]{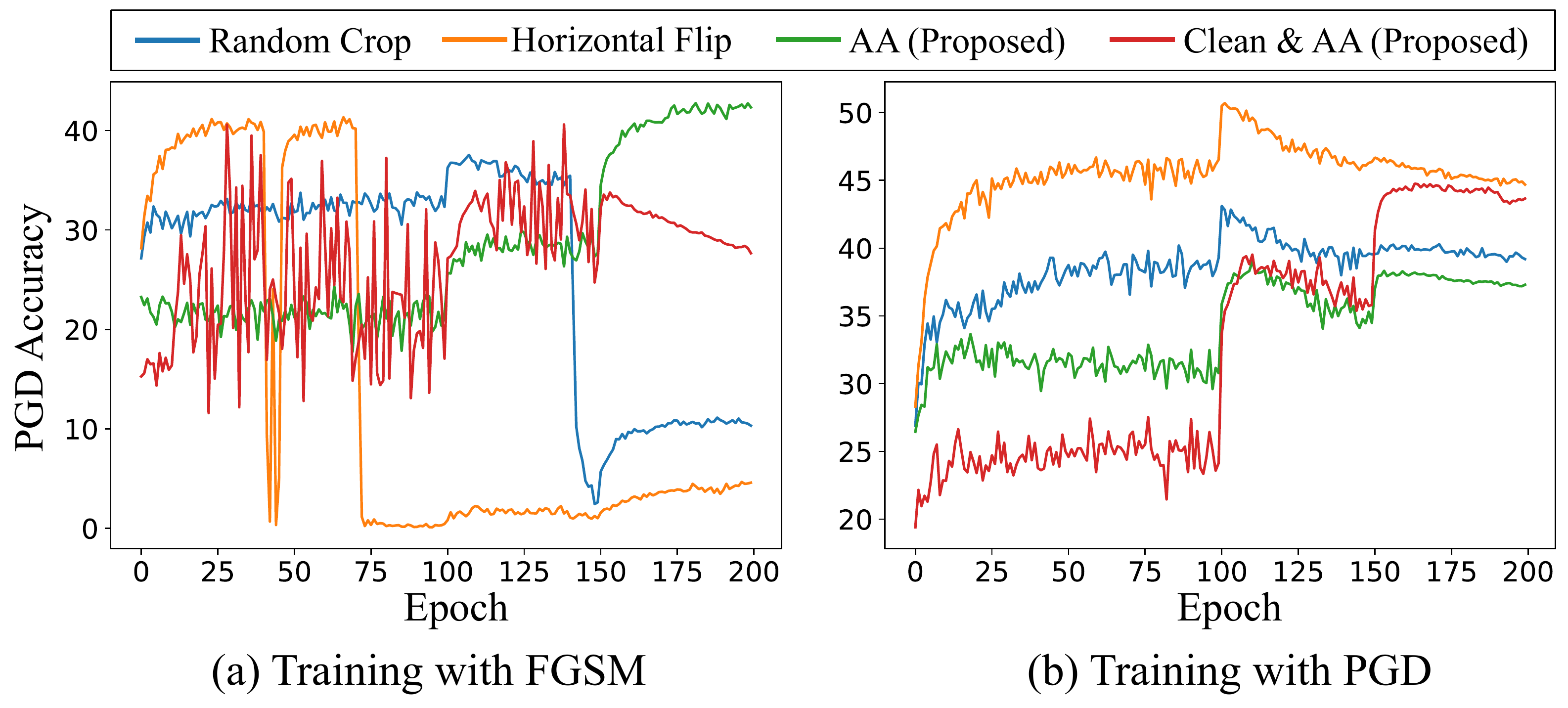}
    \caption{WideResNet-40-2 classifiers trained with conventional and proposed data augmentation methods. (a) FGSM adversarial training. (b) PGD adversarial training. }
    \label{fig:overfitting}
\end{figure}

\subsection{Prevention of Overfitting} \label{sec:experiment-overfitting}

We next show that another benefit of the proposed method is to prevent catastrophic and robust overfitting. 
Figure~\ref{fig:overfitting}(a) shows the PGD accuracy throughout the training process of FGSM adversarial training. 
As observed, the model trained with random crop and the model trained with horizontal flip suffered from catastrophic overfitting, where the PGD accuracy drastically dropped at a certain point during training. In contrast, the model trained with our method (AA) did not suffer from overfitting and could gain adversarial robustness consistently. 
Figure~\ref{fig:overfitting}(b) shows the PGD accuracy throughout the training process of PGD adversarial training. 
As observed, both the model trained with random crop and the model trained with horizontal flip suffered from robust overfitting. 
The models failed to gain adversarial robustness after the tuning of learning rates at the \mbox{100-th} epoch.
However, both the models trained with our method (AA and C\&AA) did not suffer from robust overfitting and continued to gain adversarial robustness even after the tuning of learning rates at the \mbox{100-th} and \mbox{150-th} epochs. 
Refer to the supplementary materials for more details. 

\setlength{\tabcolsep}{4pt}
\begin{table}[tb]
    \centering
    \begin{tabular}{ccccc}
        \hline
        \multicolumn{1}{c|}{}& \multicolumn{4}{c}{Accuracy (\%)} \\
        \multicolumn{1}{c|}{}&Clean&FGSM&PGD&Corr\\
        \hline
        \hline
        \multicolumn{1}{c|}{Standard} &
            94.6&\textbf{66.9}&0.0&86.0\\
        \multicolumn{1}{c|}{APR-P} &
            94.0&65.1&3.6&87.4\\
        \multicolumn{1}{c|}{$\mathcal{S}(\bm{x})$} &   
            \textbf{94.7}&\textbf{66.9}&0.1&88.5\\
        \multicolumn{1}{c|}{APR-S} &
            94.5&65.8&0.3&90.0\\
        \multicolumn{1}{c|}{APR-SP} &
            94.3&65.5&\textbf{0.9}&\textbf{90.1}\\
        \hline
    \end{tabular}
        \caption{The classification accuracy of ResNet-18 classifiers trained with different variation of APR methods. The best results are indicated in bold.}
    \label{table:ablation-aprs-augemtation}
\end{table}

\subsection{Revisiting APR}
The most similar work to ours, \citet{chen2021apr} proposed two methods that involve swapping the amplitude spectrum between images. 
\mbox{APR-P} corresponds to swap images discussed in Section~\ref{sec:fourierspecrum}.
It involves swapping the amplitude spectrum between clean and randomly-picked images. 
\mbox{APR-S} involves swapping the amplitude spectrum between two transformed images, where the image transformations used were the same as the augmentations used in AugMix~\cite{hendrycks2020augmix} (denoted by $\mathcal{S}(\bm{x})$). We revisited all the variations of APR methods by training \mbox{ResNet-50} models. Table~\ref{table:ablation-aprs-augemtation} shows the results. 
As observed, all the APR methods improved the common corruptions robustness but remained vulnerable to adversarial perturbations. 
Importantly, we noticed that $\mathcal{S}(\bm{x})$, the augmentation from AugMix itself, without any swapping of amplitude spectrum, already contributed to a big part of the improvements in common corruption robustness. 
In Table~\ref{table:ablation-apr}, we revisited the direct combination of APR methods with PGD adversarial training. 
The direct combination of APR methods and adversarial training did improve the clean~(+0.9\%) and common corruption~(+1.8\%) accuracies by a small margin with comparable adversarial robustness. 
However, the improvement is nowhere near ours (AA) in all aspects~(i.e., +6.3\% for clean, +5.7\% for PGD, and +6.8\% for common corruptions). 
Here, we want to emphasize the novelty of our method. 
The direct usage of adversarial images in the training \textit{degrades} the clean and common corruption accuracies. 
However, with AAS, we can encourage the model to learn semantic information from the phase spectrum derived from the clean images, hence significantly \textit{improve} the model performance in all aspects. Refer to the supplementary materials for more detailed comparisons. 

\setlength{\tabcolsep}{4pt}
\begin{table}[tb]
    \centering
    \begin{tabular}{ccccc}
        \hline
        \multicolumn{1}{c|}{}& \multicolumn{4}{c}{Accuracy (\%)} \\
        \multicolumn{1}{c|}{}&Clean&FGSM&PGD&Corr\\
        \hline
        \hline
        \multicolumn{1}{c|}{Adv} &   
            82.8&60.0&44.4&78.9\\
        \multicolumn{1}{c|}{APR-P + Adv} &
            65.5&47.8&44.7&62.2\\
        \multicolumn{1}{c|}{APR-S + Adv} &
            83.7&55.7&43.6&80.7\\
        \hline
        \multicolumn{1}{c|}{AA} &
        \textbf{89.1}&\textbf{74.6}&\textbf{50.1}&\textbf{85.7}\\
        \multicolumn{1}{c|}{AP} &
            87.4&72.0&46.3&83.8\\
        \hline
    \end{tabular}
    \caption{The classification accuracy of ResNet-50 classifiers trained with PGD adversarial training. The best results are indicated in bold.}
    \label{table:ablation-apr}
\end{table}

\section{Conclusion}
In this paper, we delved deeper into \textit{swap images}, which contain the original phase spectrum and the amplitude spectrum of other images. 
We demonstrated that CNNs can effectively learn the semantic information from the swap images despite having an unrelated amplitude spectrum. 
We proposed a frequency-based data augmentation, AAS, which swaps the amplitude spectrum between clean and adversarial images to generate two novel training images, i.e., adversarial amplitude~(AA) and adversarial phase~(AP) images.
AAS can be implemented in any adversarial training setup as it generates AA and AP images as substitutes for standard adversarial images. 
We demonstrated that our method significantly boosts the performance in various aspects and prevents catastrophic and robust overfitting.  
In conclusion, our method leads to CNN classifiers with general robustness against different types of perturbations and uniform performance across all types common corruptions. 
We believe that our findings are crucial in future works to train a truly robust classifier.

\end{document}


\title{\textbf{Supplementary material for} \\\textit{Exploiting Frequency Spectrum of Adversarial Images for General Robustness}}
\author{}
\date{}

\maketitle

This supplementary material presents more results of the experiments. Cross-referencing numbers here are prefixed with S (e.g., Table.~S1 or Fig.~S1). Numbers without the prefix (e.g., Table.~1 or Fig.~1) refer to numbers in the main text.

\section{Examples of swap images}
\begin{figure}[t]
    \centering
    \includegraphics[width=0.45\textwidth]{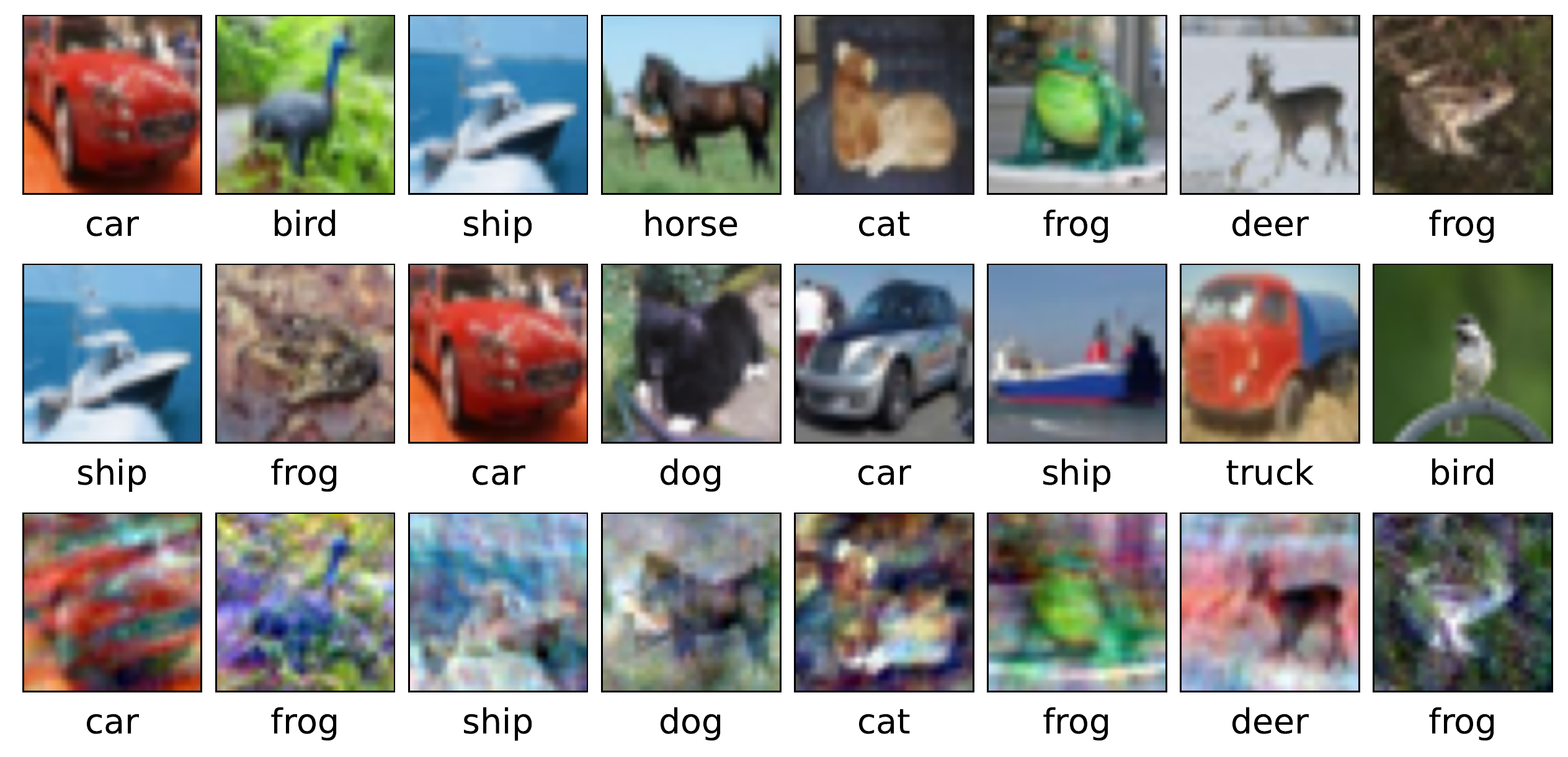}
    \caption{Examples of swap images. First row: clean training images $\bm{x}$. Second row: randomly-picked images $\bm{x}_m$. Third row: swap images $\bm{x}_\text{swap}$. The target labels are shown below each images in the first and second row. The prediction of a vanilla ResNet-18 model is shown below each image in the third row.}
    \label{fig:swap_images}
\end{figure}

Figure~\ref{fig:swap_images} shows some examples of swap images explained in Section~\swapimages. Swap images contain the phase spectrum of clean training images $\bm{x}$ and the amplitude spectrum of randomly-picked images $\bm{x}_m$. 
These images are heavily perturbed and are challenging for human to recognize. 
However, despite having the amplitude spectrum of another images, these swap images still retain the semantic information of the objects. 
As observed, a vanilla CNN model that was trained with only clean training images may make predictions based on the amplitude spectrum. For example, an image with the phase spectrum of a \textit{bird} image and the amplitude spectrum of a \textit{frog} image was classified as a \textit{frog}. 
This outcome indicates that a vanilla CNN model may not be robust against the variance in the amplitude spectrum. 
Hence, the emphasis to learn from the phase spectrum can be a key to train a robust CNN model. 
Instead of using the amplitude spectrum of another randomly-picked image, we proposed adversarial amplitude swap~(AAS) to swap the amplitude spectrum between clean and adversarial images. 
Adversarial images contain worst-case perturbations that lead to wrong predictions, hence can be used to encourage the model to learn more from the phase spectrum. 
By doing so, CNNs are allowed to exploit more information from the static phase spectrum of clean images, leading to a more generally robust classifier.

\section{Additional Results} \label{additionalresults}

In addition to the results in Section~\experiments, we present more results on the classification accuracy of models trained on CIFAR-10 and CIFAR-100 datasets~\cite{krizhevsky2009multilayersfeatures}. 
In the main paper, we demonstrated only the results of models trained with TRADES on ResNet-18~\cite{he2016resnet} models~(Table~\resnettrades). Table~\ref{table:wideresnetdensenet} shows the results of ResNet-50, WideResNet-40-2~\cite{zagoruyko2016wrn}, and DenseNet~\cite{huang2017densenet} trained on CIFAR-10 dataset with TRADES~\cite{zhang2019trades}. 
We observed the same outcomes as the results shown in Table~\resnettrades. The training with adversarial amplitude swap~(AAS) enhanced the model performances in clean, adversarial and common corruption accuracies. 
In particular, the training with adversarial amplitude~(AA) images achieved more performance gains than that of adversarial phase~(AP) images. This is because AA images contain the static phase spectrum of the clean images and stochastic amplitude spectrum of the adversarial images. 
CNNs are allowed to capture more semantic information from the static phase spectrum while learning to resist against the adversarial features in the amplitude spectrum, leading to overall improvements in performances in all aspects. 

\setlength{\tabcolsep}{4pt}
\begin{table}[t]
    \centering
    \begin{tabular}{cccccc}
        \hline
        
        \multicolumn{2}{c|}{\multirow{2}{*}{TRADES~(\mbox{$\beta=3$})}}& \multicolumn{4}{c}{Accuracy (\%)} \\
        \multicolumn{2}{c|}{}&Clean&FGSM&PGD&Corr\\
        \hline
        \hline
        \multicolumn{1}{c|}{\multirow{3}{*}{ResNet-50}}&\multicolumn{1}{c|}{Adv} &   
            81.9&56.3&44.9&78.4\\
        \multicolumn{1}{c|}{}&\multicolumn{1}{c|}{AA} &
            \textbf{86.6}&\textbf{66.9}&52.5&\textbf{83.5}\\
        \multicolumn{1}{c|}{}&\multicolumn{1}{c|}{AP} &
            86.5&65.9&\textbf{52.6}&82.5\\
        \hline
        \multicolumn{1}{c|}{\multirow{3}{*}{WideResNet}}&\multicolumn{1}{c|}{Adv} &   
            81.7&53.1&42.2&77.9\\
        \multicolumn{1}{c|}{}&\multicolumn{1}{c|}{AA} &
            \textbf{88.6}&\textbf{62.3}&\textbf{44.7}&\textbf{85.0}\\
        \multicolumn{1}{c|}{}&\multicolumn{1}{c|}{AP} &
            86.4&57.0&40.2&83.2\\
        \hline
        \multicolumn{1}{c|}{\multirow{3}{*}{DenseNet}}&\multicolumn{1}{c|}{Adv} &   
            80.1&50.5&41.9&76.7\\
        \multicolumn{1}{c|}{}&\multicolumn{1}{c|}{AA} &
            \textbf{88.3}&\textbf{51.2}&42.2&\textbf{85.1}\\
        \multicolumn{1}{c|}{}&\multicolumn{1}{c|}{AP} &
            87.2&48.6&\textbf{42.3}&84.1\\
        \hline
    \end{tabular}
        \caption{The classification accuracy of classifiers with different network architectures trained on CIFAR-10 with TRADES. The best results for each network architecture are indicated in bold.}
    \label{table:wideresnetdensenet}
\end{table}

Table~\ref{table:resnet18cifar100} shows the results of ResNet-50 models trained on CIFAR-100 with PGD adversarial training~\cite{madry2018towards}. As observed, the model trained with AA images outperformed the baseline model in all aspects~(+9.1\% on clean, +7.9\% on FGSM, +0.3\% on PGD, +8.4\% on common corruption accuracies). 
As a sum, our method, AAS, generates substitutes~(AA and AP images) for the original adversarial images in adversarial training. 
The training with these images enhances model performance in all aspects, including clean, adversarial, and common corruption accuracies.
Specifically, AA images, which contain the static phase spectrum of clean images and the stochastic amplitude spectrum of adversarial images, allow CNNs to better extract semantic information from the phase spectrum while gaining adversarial robustness.

\setlength{\tabcolsep}{6pt}
\begin{table}[t]
    \centering
    \begin{tabular}{ccccc}
        \hline
        \multicolumn{1}{c|}{ResNet-18}& \multicolumn{4}{c}{Accuracy (\%)} \\
        \multicolumn{1}{c|}{(PGD AT)}&Clean&FGSM&PGD&Corr\\
        \hline
        \hline
        \multicolumn{1}{c|}{Adv} &   
            54.4&28.7&20.8&50.4\\
        \multicolumn{1}{c|}{AA} &
            \textbf{63.5}&\textbf{36.6}&\textbf{21.1}&\textbf{58.8}\\
        \multicolumn{1}{c|}{AP} &
            60.8&35.2&20.8&56.6\\
        \hline
    \end{tabular}
        \caption{The classification accuracy of ResNet-18 trained on CIFAR-100 with PGD adversarial training~(PGD AT). The best results are indicated in bold.}
    \label{table:resnet18cifar100}
\end{table}

\setlength{\tabcolsep}{6pt}
\begin{table}[t]
    \centering
    \begin{tabular}{ccccc}
        \hline
        \multicolumn{1}{c|}{ResNet-50}& \multicolumn{4}{c}{Accuracy (\%)} \\
        \multicolumn{1}{c|}{(FGSM AT)}&Clean&FGSM&PGD&Corr\\
        \hline
        \hline
        \multicolumn{1}{c|}{Adv} &   
            \textbf{90.0}&\textbf{97.1}&1.5&84.4\\
        \multicolumn{1}{c|}{AA} &
            \textbf{90.0}&74.1&39.9&\textbf{86.2}\\
        \multicolumn{1}{c|}{AP} &
            88.2&71.8&\textbf{40.7}&84.8\\
        \hline
    \end{tabular}
        \caption{The classification accuracy of ResNet-50 trained on CIFAR-10 with FGSM adversarial training~(AT). The best results are indicated in bold.}
    \label{table:resnet50fgsm}
\end{table}

\section{Catastrophic Overfitting}

\begin{figure}
    \centering
    \includegraphics[width=0.5\textwidth]{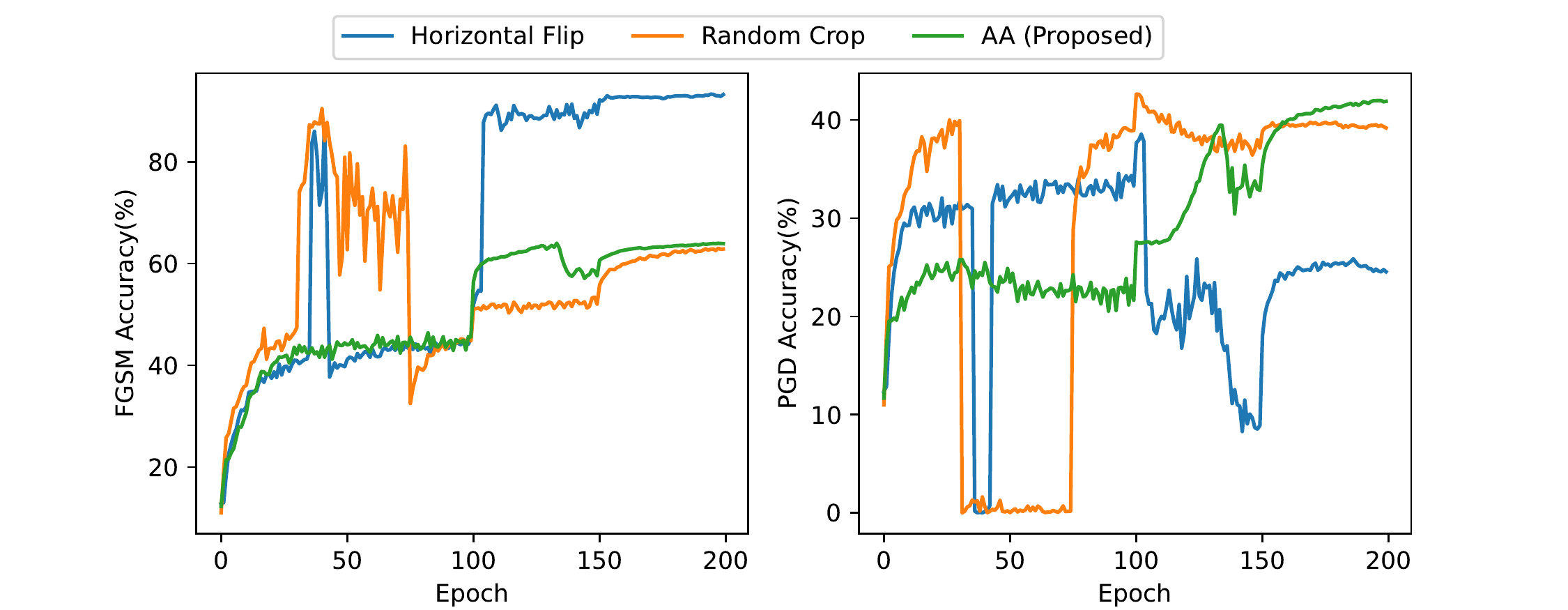}
    \caption{ResNet-50 trained with FGSM adversarial training on CIFAR-10.}
    \label{fig:catastrophic}
\end{figure}

Table~\ref{table:resnet50fgsm} shows the results of ResNet-50 models trained with FGSM adversarial training~\cite{goodfellow2015explaining}. 
FGSM adversarial training is an inexpensive training method compared to the PGD adversarial training. Models trained with FGSM adversarial training can sometimes be robust against PGD perturbations. However, the models tend to suffer from catastrophic overfitting~\cite{kim2020understandingco, kang2021understandingco}. This phenomenon can be observed in the baseline model~(Adv). The baseline model performed superbly against FGSM perturbations~(97.1\%) but was totally vulnerable against PGD perturbations~(1.5\%). 
In contrast, the models trained with AAS~(AA and AP) did not suffer from catastrophic overfitting. 
For a closer look into the effectiveness of AAS in preventing catastrophic overfitting, we showed the accuracy of ResNet-50 models, which were trained with different data augmentation method independently, on FGSM-perturbed images and PGD-perturbed images along the training process in Figure~\ref{fig:catastrophic}.
As observed, the model trained with horizonal flip suffered from catastrophic overfitting. The model trained with random crop showed improvement as the model suffered from catastrophic overfitting during the early stage of training but was able to gain PGD robustness during the later stage of training. 
In contrast, the model trained with AAS did not suffer from catastrophic overfitting and continued to gain PGD robustness along the whole training process. 
AA images contained only the amplitude spectrum of the adversarial images, leading to a slow start in gaining the adversarial robustness~(e.g., PGD robustness). 
However, as the training proceeds, the model was able to gain more robustness against the PGD perturbations than the model trained with random crop. 
In general, adversarial training requires careful tuning of the parameters to achieve good performance. 
In this paper, we focused on demonstrating the effectiveness of the proposed method by comparing the model performance with the baseline model under the same training setups. 
We demonstrated that models trained with AAS consistently showed improvements in all aspects compared to the baseline models.  

\section{Comparison with APR}
In Table~\aprpgd, we demonstrated the results of the directly combining APR methods~\cite{chen2021apr} and PGD adversarial training. We further demonstrated the results of directly combining APR methods and TRADES in Table~\ref{table:aprtrades}. 
We observed that the direct application of APR methods in adversarial training did improve the model performance on clean images and common corruptions with reasonable robustness against PGD perturbations. 
However, our method~(AA) still outperformed their methods by a significant margin. 
Although our method AAS is similar to these APR methods, there exist significant differences in our method. 
For instance, APR-S method includes the augmentation with several image transformations, such as \textit{AutoContrast}, \textit{Equalize}, \textit{Posterize}, \textit{Rotate}, \textit{Solarize}, \textit{Shear}, and \textit{Translate} for improving common corruption robustness. 
Although these image transformations do not overlap with those used in evaluation of common corruption robustness, they are not guaranteed to be complementary in enhancing adversarial robustness in adversarial training. 
In other words, these augmentations can improve model robustness against common corruptions but may reduce the gain in adversarial robustness~(APR-S+Adv). 
Our method, especially the training with AA images, aims to encourage CNN models to learn semantic information in the phase spectrum better in adversarial training setups. 
Without any help from the augmentation with image transformations, our method achieved performance gains in all aspects, leading to a generally robust classifier with uniform performance across all types of common corruptions. 
As a conclusion, our method, AAS, showed that an emphasis on the phase components in adversarial training significantly improved model performance in all aspects. 

\setlength{\tabcolsep}{4pt}
\begin{table}[t]
    \centering
    \begin{tabular}{ccccc}
        \hline
        \multicolumn{1}{c|}{TRADES}& \multicolumn{4}{c}{Accuracy (\%)} \\
        \multicolumn{1}{c|}{(\mbox{$\beta=3$})}&Clean&FGSM&PGD&Corr\\
        \hline
        \hline
        \multicolumn{1}{c|}{Adv} &   
            82.5&54.3&43.1&79.1\\
        \multicolumn{1}{c|}{APR-P + Adv} &
            85.8&55.8&33.9&83.3\\
        \multicolumn{1}{c|}{APR-S + Adv} &
            85.2&50.9&30.8&82.1\\
        \hline
        \multicolumn{1}{c|}{AA} &
            \textbf{88.0}&\textbf{67.4}&\textbf{51.8}&\textbf{84.6}\\
        \multicolumn{1}{c|}{AP} &
            85.8&66.0&50.2&82.8\\
        \hline
    \end{tabular}
    \caption{The classification accuracy of ResNet-18 classifiers trained with TRADES. The best results are indicated in bold.}
    \label{table:aprtrades}
\end{table}